\documentclass[10pt,twocolumn,letterpaper]{article}

\usepackage{cvpr}              

\usepackage{graphicx}
\usepackage{amsmath}
\usepackage{amssymb}
\usepackage{booktabs}
\usepackage{xcolor}
\usepackage[accsupp]{axessibility}
\usepackage{url}            
\usepackage{booktabs}       
\usepackage{amsfonts}       
\usepackage{nicefrac}       
\usepackage{microtype}      
\usepackage{xcolor}         

\usepackage{mathtools}
\usepackage{graphicx}
\usepackage{amsmath}
\usepackage{amssymb}
\usepackage{booktabs}
\usepackage{multirow} 
\usepackage{multicol}
\usepackage{algpseudocode}
\usepackage{algorithm}
\usepackage{caption,subcaption}
\usepackage{pgfplots}
\pgfplotsset{compat = 1.3}
\usepackage{bbm}
\usepackage{mwe,tikz}\usepackage[percent]{overpic}

\usepackage[pagebackref,breaklinks,colorlinks]{hyperref}

\usepackage[capitalize]{cleveref}
\crefname{section}{Sec.}{Secs.}
\Crefname{section}{Section}{Sections}
\Crefname{table}{Table}{Tables}
\crefname{table}{Tab.}{Tabs.}

\def\cvprPaperID{9625} 
\def\confName{CVPR}
\def\confYear{2023}

\begin{document}

\title{MarginMatch: Improving Semi-Supervised Learning with Pseudo-Margins}

\author{Tiberiu Sosea \quad\quad\quad  Cornelia Caragea \\
University of Illinois Chicago\\
{\small {\color{blue}{\tt tsosea2@uic.edu}} \quad\quad\quad {\color{blue}{\tt cornelia@uic.edu}}}
}
\maketitle

\begin{abstract}
We introduce MarginMatch, a new SSL approach combining consistency regularization and pseudo-labeling, with its main novelty arising from the use of unlabeled data training dynamics to measure pseudo-label quality. Instead of using only the model's confidence on an unlabeled example at an arbitrary iteration to decide if the example should be masked or not, MarginMatch also analyzes the {\em behavior} of the model on the pseudo-labeled examples as the training progresses, to ensure low quality predictions are masked out. MarginMatch brings substantial improvements on four vision benchmarks in low data regimes and on two large-scale datasets, emphasizing the importance of enforcing high-quality pseudo-labels. Notably, we obtain an improvement in error rate over the state-of-the-art of $3.25\%$ on CIFAR-100 with only $25$ labels per class and of $3.78\%$ on STL-10 using as few as $4$ labels per class. We make our code available at \url{https://github.com/tsosea2/MarginMatch}.
\end{abstract}

\vspace{-5mm}
\section{Introduction}
\label{sec:intro}

Deep learning models have seen tremendous success in many vision tasks \cite{10.5555/2999134.2999257,tan2019efficientnet,szegedy2015going,he2016deep,mahajan2018exploring}. This success can be attributed to their scalability, being able to produce better results when they are trained on large datasets in a supervised fashion \cite{mahajan2018exploring,tan2019efficientnet,hestness2017deep,raffel2019exploring,xie2020self,radford2019language}. Unfortunately, large labeled datasets annotated for various tasks and domains 
are difficult to acquire and demand considerable annotation effort or domain expertise. 
Semi-supervised learning (SSL) is a powerful approach that mitigates the requirement for large 
labeled datasets by effectively making use of information from unlabeled data, and thus, has been studied extensively in vision \cite{miyato2018virtual,sajjadi2016regularization,laine2016temporal,tarvainen2017mean,berthelot2019mixmatch,berthelot2019remixmatch,xie2019unsupervised,lee2013pseudo,sajjadi2016mutual,4129456,verma2021interpolation}. 

\begin{figure*} \centering
\begin{tikzpicture}
\node[text width=1.6cm] at (-5.7,4.6){{\small FixMatch}};
        \node[rectangle,minimum width = 17cm,minimum height = 2.8cm,draw] (r) at (1.8,2.2) {};
        \node[text width=17cm] at (2,4){{\small Predicted}: \enskip \enskip {\footnotesize onion} \quad \quad \enskip {\footnotesize elephant} \quad \quad {\footnotesize fossa} \quad  \enskip {\footnotesize green pepper} \quad \enskip {\footnotesize pop art} \enskip \quad \enskip {\footnotesize crowd} \quad \enskip \enskip {\footnotesize firefighter} \enskip \quad \enskip {\footnotesize horse} \quad  \quad \enskip \enskip {\footnotesize crowd} \newline {\small Actual}: \quad \enskip  {\footnotesize bell pepper} \enskip \enskip \enskip {\footnotesize camel} \enskip \enskip \quad {\footnotesize cougar} \quad \enskip \enskip \enskip {\footnotesize handrail} \quad \quad {\footnotesize poncho} \quad \enskip \enskip {\footnotesize uniform} \quad \enskip {\footnotesize voleyball} \enskip \quad \enskip {\footnotesize bison} \quad \quad \enskip {\footnotesize meat market}};
     \node (fig1) at (-4.3,5.4)
      {\includegraphics[scale=0.7]{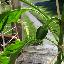}};   
     \node (fig2) at (-2.6,5.4)
      {\includegraphics[scale=0.7]{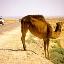}}; 
     \node (fig3) at (-0.9,5.4)
      {\includegraphics[scale=0.7]{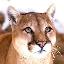}};
     \node (fig1) at (0.8,5.4)
      {\includegraphics[scale=0.7]{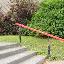}};  
     \node (fig2) at (2.5,5.4)
      {\includegraphics[scale=0.7]{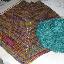}}; 
     \node (fig3) at (4.2,5.4)
      {\includegraphics[scale=0.7]{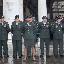}};
     \node (fig1) at (5.9,5.4)
      {\includegraphics[scale=0.7]{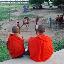}};  
     \node (fig2) at (7.6,5.4)
      {\includegraphics[scale=0.7]{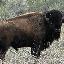}}; 
     \node (fig3) at (9.3,5.4)
      {\includegraphics[scale=0.7]{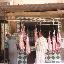}};
\node[text width=1.6cm] at (-5.8,1.8){{\small FlexMatch}};
        \node[rectangle,minimum width = 17cm,minimum height = 2.8cm,draw] (r) at (1.8,5) {};
        \node[text width=17cm] at (2,1.2){{\small Predicted}: \enskip  \enskip {\footnotesize screen} \quad \quad {\footnotesize pyramid} \enskip \quad {\footnotesize decoration} \quad \quad {\footnotesize scale} \quad \quad \enskip {\footnotesize computer} \quad \enskip \enskip {\footnotesize carpet} \quad \enskip \enskip {\footnotesize cabbage} \enskip \enskip \quad \enskip {\footnotesize tower} \quad \quad \enskip {\footnotesize screen} \newline {\small Actual}: \enskip \enskip \enskip {\footnotesize stopwatch} \enskip \enskip \enskip {\footnotesize obelisk} \quad \quad \enskip {\footnotesize socks} \quad \enskip  {\footnotesize parking meter} \enskip \enskip \enskip {\footnotesize heater} \enskip \quad {\footnotesize teddy bear} \enskip \enskip {\footnotesize cauliflower} \enskip \quad \enskip {\footnotesize torch} \quad  \quad \enskip \enskip {\footnotesize ipod}};
     \node (fig1) at (-4.3,2.7)
      {\includegraphics[scale=0.7]{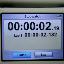}}; 
     \node (fig2) at (-2.6,2.7)
      {\includegraphics[scale=0.7]{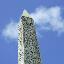}}; 
     \node (fig3) at (-0.9,2.7)
      {\includegraphics[scale=0.7]{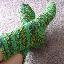}};
     \node (fig1) at (0.8,2.7)
      {\includegraphics[scale=0.7]{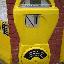}}; 
     \node (fig2) at (2.5,2.7)
      {\includegraphics[scale=0.7]{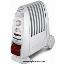}}; 
     \node (fig3) at (4.2,2.7)
      {\includegraphics[scale=0.7]{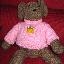}};
     \node (fig1) at (5.9,2.7)
      {\includegraphics[scale=0.7]{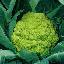}}; 
     \node (fig2) at (7.6,2.7)
      {\includegraphics[scale=0.7]{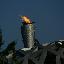}}; 
     \node (fig3) at (9.3,2.7)
      {\includegraphics[scale=0.7]{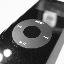}};
\end{tikzpicture}
\caption{Incorrect pseudo-labels propagated until the end of the training process for FixMatch and FlexMatch on ImageNet.}
\label{mistakes}
\end{figure*}

Recent SSL approaches integrate two important components: 
consistency regularization \cite{xie2019unsupervised,NEURIPS2021_995693c1} and pseudo-labeling \cite{lee2013pseudo}. 
Consistency regularization works on the assumption that a model should output similar predictions when fed perturbed versions of the same image, whereas pseudo-labeling uses the model's predictions of unlabeled examples as labels to train against. For example, Sohn et al. \cite{sohn2020fixmatch} introduced FixMatch that combines consistency regularization on weak and strong augmentations with pseudo-labeling.
FixMatch  relies heavily on a high-confidence threshold to compute the unsupervised loss, disregarding any pseudo-labels whose confidence falls below this threshold.
While training using only high-confidence pseudo-labels has shown to consistently reduce the confirmation bias \cite{DBLP:journals/corr/abs-1908-02983}, this rigid threshold allows access only to a small amount of unlabeled data for training, and
thus, ignores a considerable amount of 
unlabeled examples for which the model's predictions do not exceed the confidence threshold. More recently, Zhang et al. \cite{NEURIPS2021_995693c1} introduced FlexMatch that relaxes the rigid confidence threshold in FixMatch to account for the model's learning status of each class 
in that it adaptively scales down the threshold for a class to encourage the model to learn from more examples from that class. 
The flexible thresholds in FlexMatch allow the model to have access to a much larger and diverse set of unlabeled data to learn from, but lowering the thresholds can lead to the introduction of wrong pseudo-labels, 
which are extremely harmful for generalization. Interestingly, even when the high-confidence threshold is used in FixMatch can result in wrong pseudo-labels. 
See Figure \ref{mistakes} for incorrect pseudo-labels detected in the training set after we apply FixMatch and FlexMatch on ImageNet.
%
%
We posit that a drawback of FixMatch and FlexMatch and in general of any pseudo-labeling approach is that they use the confidence of the model only at the current iteration to enforce quality of pseudo-labels and completely ignore model's predictions at prior iterations.

In this paper, we propose MarginMatch, a new SSL approach that monitors the {\em behavior} of the model on the unlabeled examples as the training progresses, from the beginning of training until the current iteration, instead of using only the model's current \emph{belief} about an unlabeled example (i.e., its confidence at the current iteration) to decide if the example should be masked or not. 

We estimate a pseudo-label's contribution to learning and generalization by introducing pseudo-margins of unlabeled examples averaged across training iterations. Pseudo-margins of unlabeled examples extend the margins from machine learning \cite{DBLP:journals/corr/BartlettFT17, NEURIPS2020_c6102b37,47365,https://doi.org/10.48550/arxiv.1810.00113} which provide a measure of confidence of the outputs of the model and capture the difference between the output for the correct (gold) label and the other labels. In our case, the pseudo-margins capture how much larger the assigned logit (the logit corresponding to the argmax of the model's prediction) is compared with all other logits at iteration $t$.
Similar to FlexMatch, in MarginMatch we take advantage of the flexible confidence thresholds to allow the model to learn from larger and more diverse sets of unlabeled examples, but unlike FlexMatch, we train the model itself to identify the characteristics of mislabeled pseudo-labels simply by monitoring the model's training dynamics on unlabeled data over the iterations.

\looseness=-1
We carry out comprehensive experiments using established SSL experimental setups 
on CIFAR-10, CIFAR-100 \cite{krizhevsky2009learning}, SVHN \cite{netzer2011reading}, STL-10 \cite{coates2011analysis}, ImageNet \cite{5206848}, and WebVision \cite{DBLP:journals/corr/abs-1708-02862}. Despite its simplicity, our findings indicate that MarginMatch produces improvements in performance over strong baselines and prior works on all datasets at no additional computational cost. Notably, compared to current state-of-the-art, on CIFAR-100 we see $3.02\%$ improvement in error rate using only $4$ labels per class and $3.78\%$ improvement on STL-10 using the same extremely label-scarce setting of $4$ labels per class. In addition, on 
ImageNet \cite{5206848} and WebVision \cite{DBLP:journals/corr/abs-1708-02862} 
we find that MarginMatch pushes the state-of-the-art error rates by $0.97\%$ on ImageNet and by $0.79\%$ on WebVision.

Our contributions are as follows: 
\vspace{-1mm}
\begin{enumerate}
\item 
We introduce a new SSL approach which we call MarginMatch that enforces high pseudo-label quality during training. Our approach allows access to a large set of unlabeled data to learn from (thus, incorporating more information from unlabeled data) 
and, at the same time, monitors the training dynamics of unlabeled data as training progresses to detect and filter out potentially incorrect pseudo-labels.
\vspace{-1mm}
\item
We show that MarginMatch outperforms existing works on six well-established computer vision benchmarks showing larger improvements in error rates especially on challenging datasets, while achieving similar convergence performance (or better) than prior works.  
\vspace{-1mm}
\item We perform a comprehensive analysis of our approach and indicate potential insights into why our MarginMatch substantially outperforms other SSL techniques.
\end{enumerate}

\section{MarginMatch}
\label{sec:method}

\paragraph{Notation} 
Let $L=\{(x_{1},y_{1}),...,(x_{B},y_{B})\}$ be a batch of size $B$ of \textbf{labeled} examples 
and $U=\{\hat{x}_{1},...,\hat{x}_{\nu B}\}$ be a batch of size $\nu B$ of \textbf{unlabeled} examples, where $\nu$ is the batch-wise ratio of unlabeled to labeled examples. Let $p_{\theta}(y|x)$ denote the class distribution produced by model $\theta$ on input image $x$ and $\hat{p}_{\theta}(y|x)$ denote the argmax of this distribution as a one-hot label. Let also $H(p,q)$ denote the cross-entropy between two probability distributions $p$ and $q$.

\subsection{Background}

\noindent
Consistency regularization \cite{sajjadi2016regularization} is an important component in recent semi-supervised learning approaches and relies on the continuity assumption \cite{NIPS2014_66be31e4,laine2016temporal} that the model should output similar predictions on multiple perturbed versions of the same input $x$. As mentioned above, examples of two such approaches are FixMatch \cite{sohn2020fixmatch} and FlexMatch \cite{NEURIPS2021_995693c1} that use consistency regularization at their core combined with psedo-labeling. In psedo-labeling \cite{lee2013pseudo}, a model itself is used to assign artificial labels for unlabeled data and only artificial labels whose largest class probability is above a predefined confidence threshold are used during training. 
%

Specifically, FixMatch \cite{sohn2020fixmatch} 
predicts artificial labels for unlabeled examples using a weakly-augmented version of each unlabeled example and then employs the artificial labels as pseudo-labels to train against but this time using a strongly-augmented version of each unlabeled example. That is, FixMatch minimizes the following batch-wise consistency loss on unlabeled data:
\vspace{-2mm}
\begin{multline}
    \mathcal{L}_{u} = \sum_{i=1}^{\nu B} \mathbbm{1} (\max(p_{\theta}(y|\pi(\hat{x}_{i}))) > \tau) \quad \times \\
    H(\hat{p}_{\theta}(y|\pi(\hat{x}_{i})), p_{\theta}(y|\Pi(\hat{x}_{i})))
\label{fixmatch_vanilla}
\end{multline}
\noindent

\noindent
where $\tau$ is a confidence threshold, $\pi$ and $\Pi$ are weak and strong augmentations, respectively, and $\mathbbm{1}$ is the indicator function. 
Therefore, 
the low-confidence examples (lower than $\tau$) are completely ignored despite containing potentially useful information for model training.

\vspace{2mm}
\noindent 
FlexMatch \cite{NEURIPS2021_995693c1} argues that using a {\em fixed} threshold $\tau$ to filter the unlabeled data 
ignores the learning difficulties of different classes, and thus, introduces class-dependent thresholds, which are obtained by adaptively scaling $\tau$ depending on the learning status of each class. 
FlexMatch assumes that a class with fewer examples above the fixed threshold $\tau$ has a greater learning difficulty, and hence, it adaptively lowers the threshold $\tau$ to encourage more training examples from this class to be learned.
The learning status $\alpha_{c}$ for a class $c$ is simply computed as the number of unlabeled examples that are predicted in class $c$ and pass the fixed threshold $\tau$:

\vspace{-4mm}
\begin{equation}
    \alpha_{c} = \sum_{i=1}^{n} \mathbbm{1}(\max(p_{\theta}(y|\pi(\hat{x}_{i}))) > \tau)\mathbbm{1}(\hat{p}_{\theta}(y|\pi(\hat{x}_{i})) = c)
\label{alpha_c}
\end{equation}

\noindent
where $n$ is the total number of unlabeled examples. This learning effect is then normalized and used to obtain the class-dependent threshold for each class $c$:

\begin{equation}
    \mathcal{T}_{c} = \frac{\alpha_{c}}{\underset{c}{\max}(\alpha_{c})} \times \tau
\label{flexible_threshold}
\end{equation}

\noindent
In practice, FlexMatch iteratively computes new thresholds after each complete pass through unlabeled data, hence we can parameterize $\mathcal{T}_{c}$ as $\mathcal{T}_{c}^t$, denoting the threshold obtained at iteration $t$. 
The unlabeled loss is then obtained by plugging in the adaptive threshold $\mathcal{T}_{c}^t$ in Eq. \ref{fixmatch_vanilla}:

\vspace{-3mm}
 \begin{multline}
    \mathcal{L}_{u} = \sum_{i=1}^{\nu B} \mathbbm{1} (\max(p_{\theta}(y|\pi(\hat{x}_{i}))) > \mathcal{T}_{\hat{p}_{\theta}(y|\pi(\hat{x}_{i}))}^t) \quad \times \\ H(\hat{p}_{\theta}(y|\pi(\hat{x}_{i})), p_{\theta}(y|\Pi(\hat{x}_{i})))
\label{flexmatch_vanilla}
\end{multline}
\noindent

The aforementioned works use the confidence of the model {\em solely at the current iteration} to enforce quality of pseudo-labels. We believe this is not sufficient as it provides only a myopic view of the model's behavior (i.e., its confidence) on unlabeled data (at a single iteration) and may result in wrong pseudo-labels even when the confidence threshold is high enough (e.g., if the model is miscalibrated or overly-confident \cite{pmlr-v70-guo17a}). Figure \ref{mistakes} shows examples of images that are added to the training set with a wrong pseudo-label for both FixMatch and FlexMatch. These types of unlabeled examples, which are incorrectly pseudo-labeled and used during training are particularly harmful for deep neural networks, which can attain zero training error on any dataset, even on randomly assigned labels \cite{DBLP:journals/corr/ZhangBHRV16}, resulting in poor generalization capabilities.

\subsection{Proposed Approach: MarginMatch}

We now introduce MarginMatch, our new SSL approach that 
uses the model's training dynamics on unlabeled data to improve pseudo-label data quality.
Our approach leverages consistency regularization with weak and strong augmentations and pseudo-labeling, but instead of using only the model's current \emph{belief} (i.e., its confidence at the current iteration) to decide if an unlabeled example should be used for training or not, our MarginMatch monitors the training dynamics of unlabeled data over the iterations by investigating the {\em margins} (a measure of confidence) of the outputs of the model \cite{DBLP:journals/corr/BartlettFT17}. 
The margin of a training example is a well established metric in machine learning
\cite{DBLP:journals/corr/BartlettFT17, NEURIPS2020_c6102b37,47365,https://doi.org/10.48550/arxiv.1810.00113} that quantifies 
the difference between the logit corresponding to the assigned ground truth label and the largest other logit.

\begin{algorithm*}[t]

\begin{algorithmic}[1]

\small
\Require Labeled data $L$; unlabeled data $U$; erroneous examples $E$; maximum number of iterations $T$; number of classes $C+1$ ($C$ original \mbox{   }\mbox{   }\mbox{   }\mbox{   }\mbox{   }\mbox{   }\mbox{   }\mbox{   }\mbox{   } classes plus one virtual class of erroneous examples); $\theta$ model; $\pi$ weak augmentations; $\Pi$ strong augmentations.
\State Initialize the Average Pseudo-Margin ($APM$) threshold $\gamma^{1}$ at the first iteration to a small value (e.g., $\gamma^{1} = -\infty$).
\For{\emph{t} = $1$ to $T$}
\State Estimate learning status $\alpha_{c}$ (using Eq. \ref{alpha_c}) and calculate the class-wise flexible thresholds $\mathcal{T}^{t}_c$ (using Eq. \ref{flexible_threshold}) for each class $c$.

\While{$U$ not exhausted}
\State  Labeled batch $L_{b}=\{(x_{1},y_{1}),...,(x_{B},y_{B})\}$, unlabeled batch $U_{b}=\{\hat{x}_{1},...,\hat{x}_{\nu B}\}$, erroneous (or mislabeled) batch $E_{b}= \mbox{   }\mbox{   }\mbox{   }\mbox{   }\mbox{   }\mbox{   }\mbox{   }\mbox{   }\mbox{   }\mbox{   }\mbox{   }\{(\tilde{x}_{1},C+1),...,(\tilde{x}_{B},C+1)\}$
\For{$x \in U_b \cup E_{b} $}
\State Compute logits $z_{c}$ for each class $c$ after applying weak augmentations when $x \in U_{b}$ and strong augmentations when $x \in E_{b}$.
\State Calculate pseudo-margin $PM^{t}_{c}$ (using Eq. \ref{margin_computation}) and update Average $PM^{t}_{c}$ (using Eq. \ref{average_margin}) for each $c=1$ to $C+1$.
\EndFor

\State Minimize $\mathcal{L} = \mathcal{L}_{s} + \lambda (\mathcal{L}_{u} + \mathcal{L}_{e})$

\vspace{1mm}
\State \mbox{   }\mbox{   }\mbox{   }\mbox{   }\mbox{   } $\mathcal{L}_{s} = \frac{1}{B}\sum_{i=1}^{B} H(y_{i}, p_{\theta}(y|\pi(x_{i})))$

\State \mbox{   }\mbox{   }\mbox{   }\mbox{   }\mbox{   }
$\mathcal{L}_{u} = \sum_{i=1}^{\nu B}$ $\mathbbm{1} (\mbox{APM}^{t}_{\hat{p}_{\theta}(y|\pi(\hat{x}_{i}))}(\hat{x}_{i}) > \gamma^{t})$ $ \times 
     \mathbbm{1} (\max(p_{\theta}(y|\pi(\hat{x}_{i}))) > \mathcal{T}^{t}_{\hat{p}_{\theta}(y|\pi(\hat{x}_{i}))}) \times H(\hat{p}_{\theta}(y|\pi(\hat{x}_{i})), p_{\theta}(y|\Pi(\hat{x}_{i}))) $

\State \mbox{   }\mbox{   }\mbox{   }\mbox{   }\mbox{   }
$\mathcal{L}_{e} = \sum_{i=1}^{B}  H(C + 1, p_{\theta}(y|\Pi(\tilde{x}_{i})))$

\EndWhile
\State Update $\gamma^{t+1}$ as  the $95^{th}$ percentile erroneous sample $APM_{C+1}^{t}$.
\EndFor
\end{algorithmic}
\caption{MarginMatch}
\label{MarginMatch}

\end{algorithm*}

\looseness=-1
In our SSL formulation, we redefine the concept of margins to {\em pseudo-margins} of unlabeled examples since no ground truth labels are available for the unlabeled data. 
Let $c$ be the pseudo-label (or the argmax of the prediction, i.e., $\hat{p}_{\theta}(y|\pi(\hat{x}))$) at iteration $t$ on unlabeled example $\hat{x}$ after applying weak augmentations. 
We define the {\em pseudo-margin} (PM) of $\hat{x}$ with respect to pseudo-label $c$ at iteration $t$ as follows:

\begin{equation}
    \mbox{PM}_{c}^{t}(\hat{x}) =  z_{c} - max_{c!=i}(z_{i}) \\ 
\label{margin_computation}
\end{equation}

\looseness=-1
\noindent
where $z_{c}$ is the logit corresponding to the assigned pseudo-label $c$ and $\max_{c!=i}(z_{i})$ is the largest {\em other} logit corresponding to a label $i$ different from $c$.  To monitor the 
model's predictions on $\hat{x}$ with respect to pseudo-label $c$ from the beginning of training to iteration $t$, we average all the margins with respect to $c$ from the first iteration until $t$ and obtain the average pseudo-margin (APM) as follows:

\begin{equation}
\mbox{APM}_{c}^{t}(\hat{x}) =  \frac{1}{t} \sum_{j=1}^{t} \mbox{PM}_{c}^{j}(\hat{x})
\label{average_margin}
\end{equation}

\noindent
Here $c$ acts as the ``ground truth'' label for the APM calculation. Note that if at a prior iteration $t'$, the assigned pseudo-label is different from $c$ (say $c'$), then the APM calculation at iteration $t'$ is done with respect to $c'$ (by averaging all margins with respect to $c'$ from 1 to $t'$). In practice, we maintain a vector of pseudo-margins for all classes accumulated over the training iterations and dynamically retrieve the accumulated pseudo-margin value of the argmax class $c$ to obtain the $\mbox{APM}_{c}^{t}$ at iteration $t$.

\looseness=-1
Intuitively, if $c$ is the pseudo-label of $\hat{x}$ at iteration $t$, 
then $\mbox{PM}_{c}^{t}$ with respect to class $c$ at iteration $t$ will be positive. In contrast, if the argmax of the model prediction on $\hat{x}$ at a previous iteration $t' < t$ is different from $c$, then $PM_{c}^{t'}$ at $t'$ with respect to $c$ will be negative. Therefore, if over the iterations, the model predictions do not agree frequently with the pseudo-label $c$ from iteration $t$ and the model fluctuates significantly between iterations on the predicted label, the APM for class $c$ will have a low, likely negative value. Similarly, if the model is highly uncertain of the class of $\hat{x}$ (reflected in a high entropy of the class probability distribution), the APM for class $c$ will have a low value. These capture the characteristics of mislabeled examples or of those harmful for training.  
Motivated by these observations, MarginMatch leverages the APM of the assigned pseudo-label 
$c$ and compares it with an APM threshold to mask out pseudo-labeled examples with low APMs.
Formally, the unlabeled loss in MarginMatch is:

\vspace{-6mm}
\begin{multline}
\vspace{-2mm}
    \mathcal{L}_{u} = \sum_{i=1}^{\nu B} \mathbbm{1} (\mbox{APM}^{t}_{\hat{p}_{\theta}(y|\pi(\hat{x}_{i}))}(\hat{x}_{i}) > \gamma^{t}) \quad \times \\
    \quad \quad \quad \mathbbm{1} (\max(p_{\theta}(y|\pi(\hat{x}_{i}))) > \mathcal{T}^{t}_{\hat{p}_{\theta}(y|\pi(\hat{x}_{i}))}) \quad \times \\ H(\hat{p}_{\theta}(y|\pi(\hat{x}_{i})), p_{\theta}(y|\Pi(\hat{x}_{i}))) 
\label{aum_vanilla}
\end{multline}
\noindent

\noindent
where $\gamma^{t}$ is the APM threshold at iteration $t$, estimated as explained below, and $\mathcal{T}^{t}_{\hat{p}_{\theta}(y|\pi(\hat{x}_{i}))}$ is the flexible threshold estimated as in FlexMatch \cite{NEURIPS2021_995693c1}. 
To train our model, we adopt the best practices \cite{NEURIPS2021_995693c1,sohn2020fixmatch} and 
optimize the weighted combination of the supervised and unsupervised losses:

\begin{equation}
    \mathcal{L} = \mathcal{L}_{s} + \lambda \mathcal{L}_{u}
\label{aggregate_loss}
\end{equation}

\noindent
where the supervised loss is given by:

\vspace{-3mm}
\begin{equation}
    \mathcal{L}_{s} = \frac{1}{B}\sum_{i=1}^{B} H(y_{i}, p_{\theta}(y|\pi(x_{i})))
\label{supervised_loss}
\end{equation}


\paragraph{Average Pseudo-Margin Threshold Estimation} \quad

Inspired by Pleiss et al. \cite{NEURIPS2020_c6102b37}, we propose to estimate the average pseudo-margin threshold $\gamma^t$ by analyzing the training dynamics of a special category of unlabeled examples, which we force to be \emph{erroneous} or mislabeled examples. That is, to create the sample of \emph{erroneous} examples $E$, we randomly sample a subset of unlabeled examples from $U$ that we assign to an inexistent (or virtual) class $C+1$ at the beginning of the training process and remove them from $U$. The purpose of these erroneous examples is to mimic the training dynamics of incorrectly pseudo-labeled (unlabeled) examples and use them as proxy to estimate the cutoff of (potentially) mislabeled pseudo-labels. Since the examples in $E$ {\em should} belong to one of the $C$ original classes, assigning them to the inexistent class $C+1$ makes them by definition mislabeled (see Appendix A for additional insights into this virtual class). As with all unlabeled examples from $U$, we compute $APM^{t}_{C+1}$ for the special category of erroneous examples from $E$, but unlike the unlabeled examples from $U$, the erroneous ones from $E$ have a fixed class $C+1$. To mimic the training dynamics of unlabeled examples from $U$, we use strong augmentations to compute the loss of the erroneous examples from $E$. That is, given a batch $E_{b}$ of $B$ erroneous examples, the erroneous sample loss becomes:

\vspace{-2mm}
\begin{equation}
\label{loss_th}
   \mathcal{L}_{e} = \sum_{i=1}^{B} H(C + 1, p_{\theta}(y|\Pi(\tilde{x}_{i})))
\end{equation}

\noindent
At iteration $t$, we use the APMs of the erroneous examples to choose the APM threshold $\gamma^{t}$. We set $\gamma^{t}$ as the APM of the $95^{th}$ percentile erroneous sample. The total loss becomes:

\vspace{-3mm}
\begin{equation}
    \mathcal{L} = \mathcal{L}_{s} + \lambda (\mathcal{L}_{u} + \mathcal{L}_{e})
\label{aggregate_loss}
\end{equation}

\noindent
Our full MarginMatch algorithm is shown in Algorithm \ref{MarginMatch}.

\begin{table*}[t]
\small
\centering
\resizebox{1\textwidth}!{
\begin{tabular}{l|rrr|rrr|rrr|rrr}
\toprule
Dataset & \multicolumn{3}{c}{\textsc{CIFAR-10}} & \multicolumn{3}{c}{\textsc{CIFAR-100}} & \multicolumn{3}{c}{\textsc{SVHN}} & \multicolumn{3}{c}{\textsc{STL-10}}\\
\midrule
\#Labels/Class & \multicolumn{1}{c}{$4$} & \multicolumn{1}{c}{$25$} & \multicolumn{1}{c}{$400$} & \multicolumn{1}{c}{$4$} & \multicolumn{1}{c}{$25$} & \multicolumn{1}{c}{$100$} & \multicolumn{1}{c}{$4$} & \multicolumn{1}{c}{$25$} & \multicolumn{1}{c}{$100$} & \multicolumn{1}{c}{$4$} & \multicolumn{1}{c}{$25$} & \multicolumn{1}{c}{$100$} \\
\midrule
Pseuso-Labeling & $74.61_{0.26}$ & $46.49_{2.20}$ & $15.08_{0.19}$ & $87.45_{0.85}$ & $57.74_{0.28}$ & $36.55_{0.24}$ & $64.61_{5.60}$ & $25.21_{2.03}$ & $9.40_{0.32}$ & $74.68_{0.99}$ & $55.45_{2.43}$ & $32.64_{0.71}$ \\
UDA & $10.79_{3.75}$ & $5.32_{0.06}$ & $4.41_{0.07}$ & $48.95_{1.59}$ & $29.43_{0.21}$ & $23.87_{0.23}$ & $5.34_{4.27}$ & $4.26_{0.39}$ & $1.95_{0.01}$ & $37.82_{8.44}$ & $9.81_{1.15}$ & $6.81_{0.17}$ \\
MixMatch & $45.24_{2.15}$ & $12.76_{1.14}$ & $7.13_{0.34}$ & $62.15_{2.17}$ & $41.51_{1.19}$ & $28.16_{0.24}$ & $46.18_{1.78}$ & $3.98_{0.17}$ & $3.5_{0.13}$ & $34.15_{1.54}$ & $8.95_{0.32}$ & $10.41_{0.73}$ \\
ReMixMatch & $5.27_{0.19}$ & $4.85_{0.13}$ & $4.04_{0.12}$ & $47.15_{0.76}$ & $27.14_{0.23}$ & $23.78_{0.12}$ & $4.23_{0.31}$ & $3.18_{0.04}$ & $1.94_{0.06}$ & $31.51_{0.75}$ & $8.54_{0.48}$ & $6.19_{0.24}$ \\
FixMatch & $7.8_{0.28}$ & $4.91_{0.05}$ & $4.25_{0.08}$ & $48.21_{0.82}$ & $29.45_{0.16}$ & $22.89_{0.12}$ & $3.97_{1.18}$ & {\color{blue}{$\boldsymbol{3.13_{1.03}}$}} & $1.97_{0.03}$ & $38.43_{4.14}$ & $10.45_{1.04}$ & $6.43_{0.33}$ \\
FlexMatch & $5.04_{0.06}$ & $5.04_{0.09}$ & $4.19_{0.01}$ & $39.99_{1.62}$ & $26.96_{0.08}$ & $22.44_{0.15}$ & $8.19_{3.20}$ & $7.78_{2.55}$ & $6.72_{0.30}$ & $29.15_{1.32}$ & $8.23_{0.13}$ & $5.77_{0.12}$ \\
\midrule
MarginMatch & {\color{blue}{$\boldsymbol{4.91_{0.07}}$}} & {\color{blue}{$\boldsymbol{4.73_{0.12}}$}} & {\color{blue}{$\boldsymbol{3.98_{0.02}}$}} & {\color{blue}{$\boldsymbol{36.97_{1.32}}$}} & {\color{blue}{$\boldsymbol{23.71_{0.13}}$}} & {\color{blue}{$\boldsymbol{21.39_{0.12}}$}} & {\color{blue}{$\boldsymbol{3.75_{1.20}}$}} & $3.14_{1.17}$ & {\color{blue}{$\boldsymbol{1.93_{0.01}}$}} & {\color{blue}{$\boldsymbol{25.37_{3.58}}$}} & {\color{blue}{$\boldsymbol{7.31_{0.35}}$}} & {\color{blue}{$\boldsymbol{5.52_{0.15}}$}} \\
\bottomrule
\end{tabular}
}
\caption{Test error rates on CIFAR-10, CIFAR-100, SVHN, and STL-10 datasets. Best results are shown in {\color{blue}{\bf blue}}.}
\label{tab:all_results}
\end{table*}

\paragraph{Exponential Moving Average of Pseudo-Margins} 
The current definition of APM weighs the pseudo-margin at iteration $t$ identical to the pseudo-margin at a much earlier iteration $p$ ($t >> p$). This is problematic since very old pseudo-margins eventually become deprecated (especially due to the large number of iterations through unlabeled data in consistency training ($\sim9K$)), and hence, the old margins are no longer indicative of the current learning status of the model. To this end, instead of averaging all pseudo-margins (from the beginning of training to the current iteration), we propose to use an exponential moving average to place more importance on recent iterations. Formally, APM becomes:
\vspace{-1mm}
\begin{equation} 
    \mbox{APM}^{t}_{c}(\hat{x}) =  \mbox{PM}_{c}^{t}(\hat{x}) * \frac{\delta}{1 + t} + \mbox{APM}^{t-1}_{c}(\hat{x}) * (1 - \frac{\delta}{1 + t})
\label{aum_computation}
\end{equation}

\noindent
We set the smoothing parameter $\delta$ to $0.997$ in experiments.

\section{Experiments}
\label{sec:experiments}

\noindent
\looseness=-1
We evaluate the performance of our MarginMatch on a wide range of SSL benchmark datasets. Specifically, we perform experiments with various numbers of labeled examples on CIFAR-10, CIFAR-100 \cite{krizhevsky2009learning}, SVHN \cite{netzer2011reading}, STL-10 \cite{coates2011analysis}, ImageNet \cite{5206848}, and WebVision \cite{DBLP:journals/corr/abs-1708-02862}. 
For smaller scale datasets such as CIFAR-10, CIFAR-100, SVHN, and STL-10 we randomly sample a small number of labeled examples per class (ranging from $4$ labels per class to $400$ labels per class) and treat them as labeled data, whereas the remaining labeled examples are treated as unlabeled data, except for STL-10 \cite{coates2011analysis}, which provides its own set of unlabeled examples. On ImageNet and WebVision, we use $\sim10\%$ of the available labeled examples as labeled data, with the remaining ($90\%$) being treated as unlabeled data. In all our experiments, we sample $5\%$ of the unlabeled data and place it in the set of erroneous examples.

\looseness=-1
 We report the mean and standard deviation of error rates from five runs with different parameter initializations. Similar to FixMatch \cite{sohn2020fixmatch}, we use Wide Residual Networks \cite{zagoruyko2017wide}: WRN-28-2 for CIFAR-10 and SVHN; WRN-28-8 for CIFAR-100; and WRN-37-2 for STL-10. We use ResNet-50 \cite{he2016deep} for both ImageNet and WebVision.

In our experiments, we adopt the same hyperparameters as FixMatch \cite{sohn2020fixmatch}. Specifically, we use stochastic gradient descent (SGD) with a momentum of $0.9$. We start with a learning rate of $0.03$ and employ a cosine learning rate; at iteration $k$, our learning rate is $\eta(k) = cos(\frac{7k\pi}{16K})$, where $K$ is the maximum number of iterations and is set to $2^{20}$. We also leverage the same data augmentations as in FlexMatch \cite{NEURIPS2021_995693c1}. Specifically, for weak augmentations we employ a standard flip-and-shift augmentation and use RandAugment \cite{cubuk2020randaugment} for strong augmentations. We set the batch size $B=64$,  ratio of unlabeled to labeled data in a batch to $\nu=7$ and weigh the supervised and unsupervised losses equally (i.e., $\lambda = 1$). We set our initial confidence threshold to $\tau=0.95$ and our average pseudo-margin (APM) threshold to the APM of the $95^{th}$ percentile threshold sample. To report the error rates, we compare all the approaches using the model at the end of training as in FixMatch \cite{sohn2020fixmatch}.

\begin{figure*}
\centering
\begin{tikzpicture}[scale=0.5]
\begin{axis}[
        title=CIFAR-10,
        xlabel=$\#Epochs$,
        ylabel=$Error Rate(\%)$,
        xmin=0, xmax=9000,
        ymin=0, ymax=87,
        xtick={0,1000,...,9000},
        xticklabels={0, 1K, 2K, 3K, 4K, 5K, 6K, 7K, 8K,9K},
        ytick={10,20,30,40,50,60,70,80,90,100},
        every axis plot/.append style={ultra thick}
        ]
\addplot[smooth,mark=*,blue,dashed] plot coordinates {
    (0,87)
    (1000,81)
    (2000,54)
    (3000,32)
    (4000,15)
    (5000,6)
    (6000,5)
    (7000,4)
    (8000,4)
    (9000,4)
};
\addlegendentry{FixMatch}

\addplot[smooth,mark=x,red] plot coordinates {
    (0,87)
    (1000,35)
    (2000,12)
    (3000,6)
    (4000,5)
    (5000,4)
    (6000,4)
    (7000,4)
    (8000,4)
    (9000,4)
};
\addlegendentry{FlexMatch}

\addplot[smooth,mark=x,black,dash pattern=on 1pt off 3pt on 3pt off 3pt,] plot coordinates {
    (0,87)
    (1000,37)
    (2000,21)
    (3000,5)
    (4000,4.5)
    (5000,4.5)
    (6000,5)
    (7000,5)
    (8000,4)
    (9000,4)
};
\addlegendentry{MarginMatch}
dash pattern=on 1pt off 3pt on 3pt off 3pt,
\end{axis}
\end{tikzpicture}
\begin{tikzpicture}[scale=0.5]
\begin{axis}[
        title=CIFAR-100,
        xlabel=$\#Epochs$,
        ylabel=$Error Rate(\%)$,
        xmin=0, xmax=9000,
        ymin=30, ymax=100,
        xtick={0,1000,...,9000},
        xticklabels={0, 1K, 2K, 3K, 4K, 5K, 6K, 7K, 8K,9K},
        ytick={10,20,30,40,50,60,70,80,90,100},
        every axis plot/.append style={ultra thick}
        ]
\addplot[smooth,mark=*,blue,dashed] plot coordinates {
    (0,98)
    (1000,95)
    (2000,82)
    (3000,75)
    (4000,48)
    (5000,47)
    (6000,45)
    (7000,46)
    (8000,43)
    (9000,42)
};
\addlegendentry{FixMatch}

\addplot[smooth,mark=x,red] plot coordinates {
    (0,96)
    (1000,67)
    (2000,53)
    (3000,49)
    (4000,45)
    (5000,41)
    (6000,39)
    (7000,39)
    (8000,39.5)
    (9000,39)
};
\addlegendentry{FlexMatch}

\addplot[smooth,mark=x,black,dash pattern=on 1pt off 3pt on 3pt off 3pt,] plot coordinates {
    (0,97)
    (1000,57)
    (2000,39)
    (3000,38)
    (4000,37)
    (5000,37)
    (6000,37)
    (7000,37)
    (8000,36.5)
    (9000,36)
};
\addlegendentry{MarginMatch}
dash pattern=on 1pt off 3pt on 3pt off 3pt,
\end{axis}
\end{tikzpicture}
\begin{tikzpicture}[scale=0.5]
\begin{axis}[
        title=SVHN,
        xlabel=$\#Epochs$,
        ylabel=$Error Rate(\%)$,
        xmin=0, xmax=9000,
        ymin=0, ymax=70,
        xtick={0,1000,...,9000},
        xticklabels={0, 1K, 2K, 3K, 4K, 5K, 6K, 7K, 8K,9K},
        ytick={10,20,30,40,50,60,70,80,90,100},
        every axis plot/.append style={ultra thick}
        ]
\addplot[smooth,mark=*,blue,dashed,line width=0.25mm] plot coordinates {
    (0, 71)
    (1000,65)
    (2000,58)
    (3000,52)
    (4000,35)
    (5000,25)
    (6000,10)
    (7000,7)
    (8000,6)
    (9000,4)
};
\addlegendentry{FixMatch}

\addplot[smooth,mark=x,red] plot coordinates {
    (0,70)
    (1000,32)
    (2000,12)
    (3000,10)
    (4000,7)
    (5000,7.2)
    (6000,7)
    (7000,7.2)
    (8000,7.6)
    (9000,7.5)
};
\addlegendentry{FlexMatch}

\addplot[smooth,mark=x,black,dash pattern=on 1pt off 3pt on 3pt off 3pt,] plot coordinates {
    (0,64)
    (1000,31)
    (2000,13)
    (3000,9)
    (4000,7)
    (5000,4)
    (6000,4)
    (7000,4)
    (8000,4)
    (9000,4)
};
\addlegendentry{MarginMatch}
dash pattern=on 1pt off 3pt on 3pt off 3pt,
\end{axis}
\end{tikzpicture}
\begin{tikzpicture}[scale=0.5]
\begin{axis}[
        title=STL-10,
        xlabel=$\#Epochs$,
        ylabel=$Error Rate(\%)$,
        xmin=0, xmax=9000,
        ymin=25, ymax=100,
        xtick={0,1000,...,9000},
        xticklabels={0, 1K, 2K, 3K, 4K, 5K, 6K, 7K, 8K,9K},
        ytick={10,20,30,40,50,60,70,80,90,100},
        every axis plot/.append style={ultra thick}
        ]
\addplot[smooth,mark=*,blue,dashed] plot coordinates {
    (0,86)
    (1000,81)
    (2000,76)
    (3000,57)
    (4000,49)
    (5000,46)
    (6000,41)
    (7000,39)
    (8000,38)
    (9000,37)
};
\addlegendentry{FixMatch}

\addplot[smooth,mark=x,red] plot coordinates {
    (0,87)
    (1000,64)
    (2000,46)
    (3000,40)
    (4000,39)
    (5000,37)
    (6000,36)
    (7000,35.27)
    (8000,33.18)
    (9000,32.15)
};
\addlegendentry{FlexMatch}

\addplot[smooth,mark=x,black,dash pattern=on 1pt off 3pt on 3pt off 3pt] plot coordinates {
    (0,88)
    (1000,69)
    (2000,58)
    (3000,53)
    (4000,45)
    (5000,41)
    (6000,37)
    (7000,32)
    (8000,29)
    (9000,28)
};
\addlegendentry{MarginMatch}
dash pattern=on 1pt off 3pt on 3pt off 3pt,
\end{axis}
\end{tikzpicture}
\caption{Convergence speed of MarginMatch against FixMatch and FlexMatch 
with 4 labels per class.}
\label{convergence_speed}
\end{figure*}
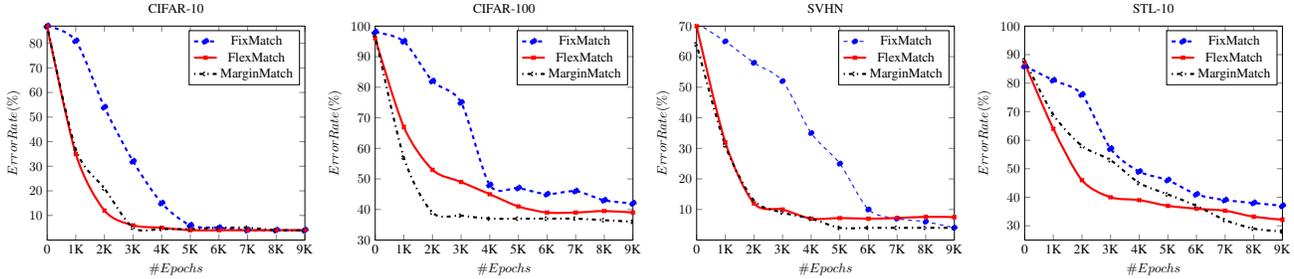

\subsection{CIFAR-10, CIFAR-100, SVHN, and STL-10}
\noindent
We compare MarginMatch against strong baselines and prior works: 
Pseudo-Labeling \cite{DBLP:journals/corr/abs-1908-02983}, Unsupervised Data Augmentation (UDA) \cite{xie2019unsupervised}, MixMatch \cite{berthelot2019mixmatch}, ReMixMatch \cite{berthelot2019remixmatch}, FixMatch \cite{sohn2020fixmatch}, and FlexMatch \cite{NEURIPS2021_995693c1}. We show in Table \ref{tab:all_results} the error rates obtained by our MarginMatch and the baselines on the CIFAR-10, CIFAR-100, SVHN and STL-10 datasets. First, we observe that our approach improves the performance on both CIFAR-10 and CIFAR-100. On CIFAR-10, MarginMatch improves performance in all data regimes upon FlexMatch \cite{NEURIPS2021_995693c1}, which is the current state-of-the-art, while mantaining a good error rate standard deviation. 
On CIFAR-100, which is significantly more challenging than CIFAR-10, we observe that MarginMatch bring substantially larger improvements. Notably, we see $3.02\%$ improvement over FlexMatch in error rate using only $4$ labels per class, and $3.25\%$ improvement using $25$ examples per class. These results on CIFAR-100 emphasize the effectiveness of MarginMatch, which performs well on a more challenging dataset.

\looseness=-1
On SVHN, our approach performs better than FixMatch using $4$ labels per class and performs similarly with FixMatch using $25$ and $100$ labels per class. However, on this dataset, MarginMatch performs much better compared with FlexMatch. For example, MarginMatch achieves $3.75\%$ error rate using $4$ labels per class, whereas FlexMatch obtains an error rate of $8.19\%$ with the same labels per class, yielding an improvement of MarginMatch of $4.44\%$ over FlexMatch. We hypothesize that the low performance of FlexMatch is due to its limitation in handling unbalanced class distributions \cite{NEURIPS2021_995693c1}. 
On STL-10, MarginMatch as well outperforms all the other approaches both in error rates and error rate standard deviation. Notably, on this dataset, our approach pushes the performance of FlexMatch by $3.78\%$ in error rate using only $4$ labels per class and by $0.92\%$ using $25$ labels per class.

Next, we compare MarginMatch with FixMatch and FlexMatch in terms of convergence speed in the extremely label-scarce setting of 4 labels per class and show these results in Figure \ref{convergence_speed}. Notably, we observe that MarginMatch has a similar convergence speed (or even better on CIFAR-100) compared with FlexMatch while achieving a lower test error rate than FlexMatch on all datasets with 4 labels per class (see Table \ref{tab:all_results}). Even more strikingly, compared with FixMatch, MarginMatch has a much superior convergence speed for a much better test error rate with 4 labels per class. This is because the rigid thresholds in FixMatch allow access only to a small amount of unlabeled data for training at each iteration and it takes a lot longer for the model to train. 

\begin{table}
\small
\centering
\resizebox{0.8\columnwidth}!{
\begin{tabular}{l|cc|cc}
\toprule
Dataset & \multicolumn{2}{c}{ImageNet} & \multicolumn{2}{c}{WebVision} \\
& \textsc{top-1} & \textsc{top-5} & \textsc{top-1} & \textsc{top-5} \\
\midrule
Supervised & $48.39$ & $25.49$ & $49.58$ & $26.78$ \\
FixMatch & $43.66$ & $21.80$ & $44.76$ & $22.65$ \\
FlexMatch & $42.02$ & $19.49$ & $43.87$ & $22.07$ \\
\midrule
MarginMatch & {\color{blue}{$\boldsymbol{41.05}$}} &
{\color{blue}{$\boldsymbol{18.28}$}} & {\color{blue}{$\boldsymbol{43.08}$}} & {\color{blue}{$\boldsymbol{21.13}$}}  \\
\bottomrule
\end{tabular}
}
\caption{Test error rates on the ImageNet and WebVision datasets. Best results are shown in {\color{blue}{\bf blue}}.}
\label{large_scale_experiment}
\vspace{-5mm}
\end{table}

\subsection{ImageNet and WebVision}
\noindent
\looseness=-1
To showcase the effectiveness of our approach in a large-scale setup, we test our MarginMatch on ImageNet \cite{5206848} and WebVision \cite{DBLP:journals/corr/abs-1708-02862} using $10\%$ labeled examples in total. We show the results obtained in Table \ref{large_scale_experiment}. We observe that our MarginMatch outperforms FixMatch and FlexMatch on both datasets. 
It is worth noting that large-scale self-supervised approaches such as SimCLR \cite{chen2020simple} achieve high performance on ImageNet but at a much higher computational cost. MarginMatch outperforms other SSL methods using the same ResNet-50 architecture at the same computational cost. We emphasize MarginMatch is most successful and relevant in low data regimes on smaller datasets.

\section{Ablation Study}
\label{sec:analysis}

\vspace{1mm}

\noindent
\looseness=-1
\textbf{Exponential Moving Average Smoothing for APM Computation} In our approach, we employ an exponential moving average (EMA) of the pseudo-margin values with a smoothing value of $\delta=0.997$ to compute the APM. We now analyze how our approach performs with different EMA smoothing values or with no EMA at all. Table \ref{smoothing} shows these results on CIFAR-100 with 4 labels per class. First, we observe that employing a simple average of pseudo-margin values for the APM computation (i.e., $\delta=1$) performs extremely poorly, obtaining a $39.72$\% error rate. This result emphasizes that margins eventually become deprecated and it is essential to scale them down in time. Using a low smoothing factor of $\delta=0.95$ is not effective either, denoting that abruptly forgetting margin values does not work either. Our chosen $\delta=0.997$ strikes a balance between the two by eliminating the harmful effects of very old margins while keeping track of a good amount of previous estimates (e.g., a margin value computed $200$ epochs previously is scaled down by $0.55$, while a margin value computed $1000$ epochs previously is scaled by $0.05$).

\begin{table}
\centering
\resizebox{0.9\columnwidth}!{
\small
\begin{tabular}{c|cccccc}
\toprule
$\delta$ & $0.95$ & $0.99$ & $0.995$ & $0.997$ & $0.999$ & $1$ \\
\midrule
\textsc{err rate} & $38.13$ & $38.05$ & $37.92$ & {\color{blue} $\bf 37.91$} & $39.12$ & $39.72$  \\
\bottomrule
\end{tabular}
}
\caption{Error rates obtained on CIFAR-100 with four examples per class and various smoothing values $\delta$. Best result is in {\color{blue}{\bf blue}}.}
\label{smoothing}
\vspace{-4mm}
\end{table}

\begin{table*}[t]
\small
\centering
\resizebox{1\textwidth}!{
\begin{tabular}{l|rrr|rrr|rrr|rrr}
\toprule
Dataset & \multicolumn{3}{c}{\textsc{CIFAR-10}} & \multicolumn{3}{c}{\textsc{CIFAR-100}} & \multicolumn{3}{c}{\textsc{SVHN}} & \multicolumn{3}{c}{\textsc{STL-10}} \\
\midrule
\#Labels/Class & \multicolumn{1}{c}{$4$} & \multicolumn{1}{c}{$25$} & \multicolumn{1}{c}{$400$} & \multicolumn{1}{c}{$4$} & \multicolumn{1}{c}{$25$} & \multicolumn{1}{c}{$100$} & \multicolumn{1}{c}{$4$} & \multicolumn{1}{c}{$25$} & \multicolumn{1}{c}{$100$} & \multicolumn{1}{c}{$4$} & \multicolumn{1}{c}{$25$} & \multicolumn{1}{c}{$100$} \\
\midrule
Avg Confidence & $23.87_{2.73}$ & $14.21_{1.37}$ & $7.54_{0.78}$ & $41.23_{2.15}$ & $31.49_{1.48}$ & $24.11_{2.36}$ & $8.99_{4.27}$ & $6.54_{0.39}$ & $4.73_{0.01}$ & $31.67_{8.44}$ & $14.87_{1.15}$ & $7.59_{0.17}$ \\
Avg Entropy & $8.58_{0.41}$ & $6.18_{0.15}$ & $5.85_{0.12}$ & $45.10_{0.91}$ & $26.02_{1.11}$ & $22.13_{0.25}$ & $15.69_{1.25}$ & $12.74_{0.78}$ & $9.33_{0.05}$ & $29.54_{3.51}$ & $10.63_{1.35}$ & $10.84_{0.47}$ \\
Avg Margin & $7.25_{0.29}$ & $5.38_{0.76}$ & $4.73_{0.09}$ & $39.72_{1.52}$ & $25.21_{0.52}$ & $23.18_{0.17}$ & $18.45_{1.36}$ & $11.29_{0.93}$ & $8.40_{0.04}$ & $28.45_{4.28}$ & $9.34_{1.34}$ & $7.59_{0.21}$ \\
\midrule
\midrule
EMA Confidence &  ${4.91_{0.45}}$ & $4.74_{0.09}$ & $3.99_{0.06}$ & $38.67_{0.74}$ & $25.61_{0.12}$ & $21.48_{0.17}$  & $3.84_{0.23}$ & $3.25_{0.03}$ & $1.93_{0.09}$ & $25.9_{0.81}$ & $7.6_{0.42}$ & $5.74_{0.57}$ \\
EMA Entropy & $6.4_{0.43}$ & $8.34_{0.12}$ & $4.21_{0.09}$ & $41.63_{0.76}$ & $36.84_{0.13}$ & $22.52_{0.07}$   & $3.81_{1.26}$ & $3.17_{0.87}$ & $2.14_{0.04}$ & $27.21_{4.05}$ & $8.28_{1.01}$ & $6.79_{0.27}$ \\
EMA Margin & {\color{blue}{$\boldsymbol{4.91_{0.07}}$}} & {\color{blue}{$\boldsymbol{4.73_{0.12}}$}} & {\color{blue}{$\boldsymbol{3.98_{0.02}}$}} & {\color{blue}{$\boldsymbol{36.97_{1.32}}$}} & {\color{blue}{$\boldsymbol{23.71_{0.13}}$}} & {\color{blue}{$\boldsymbol{21.39_{0.12}}$}} & {\color{blue}{$\boldsymbol{3.75_{1.20}}$}} & {\color{blue}{$\boldsymbol{3.14_{1.17}}$}} & {\color{blue}{$\boldsymbol{1.93_{0.01}}$}} & {\color{blue}{$\boldsymbol{25.37_{3.58}}$}} & {\color{blue}{$\boldsymbol{7.31_{0.35}}$}} & {\color{blue}{$\boldsymbol{5.52_{0.15}}$}} \\
\bottomrule
\end{tabular}
}
\caption{Test error rates comparing pseudo-margin with confidence and entropy. Best results are shown in {\color{blue}{\bf blue}}.}
\label{tab:ablated}
\vspace{-4mm}
\end{table*}

\vspace{2mm}
\noindent
\looseness=-1
\textbf{Pseudo-Margin vs. Other Measures for Pseudo-Label Correctness} \quad
Our MarginMatch monitors the pseudo-margins of a model's predictions across training iterations to ensure the quality of pseudo-labels. However, other measures such as confidence or entropy exist that can assess the pseudo-label correctness. Hence, we perform an ablation where we replace the pseudo-margins in our MarginMatch with average confidence and entropy and compare their performance. 
Specifically, we design the following approaches: \textbf{1) Avg Confidence} monitors the confidence of the prediction for each unlabeled example and takes the average over the training iterations; 
\textbf{2) Avg Entropy} monitors the entropy of the class probability distribution for each unlabeled example and takes the average across the training iterations. In addition, we also consider \textbf{3) EMA Confidence} and \textbf{4) EMA Entropy} which are similar to Avg Confidence and Avg Entropy, respectively, but use an exponential moving average (EMA) instead of the simple averaging.
The estimation of the threshold for each of these approaches is done in a similar manner as for pseudo-margins, using erroneous samples and considering the value of the 95th percentile erroneous sample as the threshold. 

We show the results obtained using these approaches in Table \ref{tab:ablated}. First, we observe that all measures (pseudo-margin, confidence and entropy) with EMA perform better than their counterpart with simple averaging. Second, EMA Margin achieves the lowest test error rates compared with EMA Confidence and EMA Entropy. Thus, we conclude that pseudo-margins provide an excellent measure for pseudo-label correctness. See Appendix B for some additional insights into why EMA Margin outperforms EMA confidence and entropy.

\begin{figure}
\centering
\begin{tikzpicture}[scale=0.5]
\begin{axis}[
        title=CIFAR-100,
        xlabel=$\#Epochs$,
        ylabel=$MaskRate(\%)$,
        xmin=4500, xmax=9000,
        ymin=15, ymax=30,
        xtick={4500,5000,...,9000},
        xticklabels={4.5K, 5K, 5.5K, 6K, 6.5K, 7K, 7.5K, 8K, 8.5K},
        ytick={10,20,30,40,50,60,70,80,90,100},
        every axis plot/.append style={ultra thick}
        ]
\addplot[smooth,mark=*,blue,dashed] plot coordinates {
    (4500,25.3)
    (5000,23.7)
    (5500,23.1)
    (6000,23.4)
    (6500,23.6)
    (7000,23.3)
    (7500,22.2)
    (8000,22.1)
    (8500,22.2)
    (9000,22.3)
};
\addlegendentry{FixMatch}

\addplot[smooth,mark=x,red] plot coordinates {
    (4500,20)
    (5000,20.1)
    (5500,20.0)
    (6000,19.5)
    (6500,19.2)
    (7000,18.1)
    (7500,17.3)
    (8000,17.1)
    (8500,17.03)
    (9000,17)
};
\addlegendentry{FlexMatch}

\addplot[smooth,mark=x,black,dash pattern=on 1pt off 3pt on 3pt off 3pt,] plot coordinates {
    (4500,21.5)
    (5000,21.4)
    (5500,20.6)
    (6000,20.5)
    (6500,20.9)
    (7000,20.6)
    (7500,20.1)
    (8000,19.4)
    (8500,18.3)
    (9000,18.01)
};
\addlegendentry{MarginMatch}
dash pattern=on 1pt off 3pt on 3pt off 3pt,
\end{axis}
\end{tikzpicture}
\begin{tikzpicture}[scale=0.5]
\begin{axis}[
        title=CIFAR-100,
        xlabel=$\#Epochs$,
        ylabel=$Impurity(\%)$,
        xmin=4500, xmax=9000,
        ymin=11, ymax=20,
        xtick={4500,5000,...,9000},
        xticklabels={4.5K, 5K, 5.5K, 6K, 6.5K, 7K, 7.5K, 8K, 8.5K},
        ytick={10, 11, 12, 13, 14, 15, 16, 17, 18, 19, 20},
        every axis plot/.append style={ultra thick}
        ]
\addplot[smooth,mark=*,blue,dashed] plot coordinates {
    (4500,15.4)
    (5000,15.3)
    (5500,15.5)
    (6000,14.7)
    (6500,14.6)
    (7000,14.1)
    (7500,13.5)
    (8000,13.3)
    (8500,13.3)
    (9000,13.2)
};
\addlegendentry{FixMatch}

\addplot[smooth,mark=x,red] plot coordinates {
    (4500,16.5)
    (5000,15.7)
    (5500,15.9)
    (6000,15.2)
    (6500,14.7)
    (7000,14.6)
    (7500,14.4)
    (8000,14.3)
    (8500,14.2)
    (9000,14.1)
};
\addlegendentry{FlexMatch}

\addplot[smooth,mark=x,black,dash pattern=on 1pt off 3pt on 3pt off 3pt,] plot coordinates {
    (4500,15.2)
    (5000,14.7)
    (5500,14.8)
    (6000,14.5)
    (6500,14.6)
    (7000,13.5)
    (7500,13.3)
    (8000,13.2)
    (8500,13.1)
    (9000,13.0)
};
\addlegendentry{MarginMatch}
dash pattern=on 1pt off 3pt on 3pt off 3pt,
\end{axis}
\end{tikzpicture}
\caption{Mask rate and impurity on CIFAR-100 with $4$ labeled examples per class.}
\label{mask_rate_impurity}
\vspace{-6mm}
\end{figure}
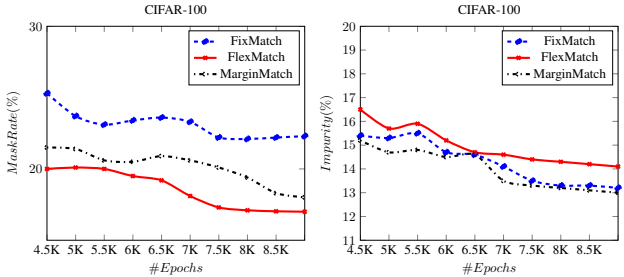

\section{Analysis}

\subsection{Mask Rate and Impurity}
\label{sec:analysis}

\noindent 
We now contrast MarginMatch with FixMatch and FlexMatch in terms of the quality of pseudo-labels using two metrics: {\em mask rate} and {\em impurity} and show these results in Figure \ref{mask_rate_impurity}, respectively, using CIFAR-100 with $4$ labels per class. {\em Mask rate} is defined as the fraction of pseudo-labeled examples that {\em do not} participate in the training at epoch t due to 
confidence masking or pseudo-margin masking (or both). 
{\em Impurity} in contrast is defined as the fraction of pseudo-labeled examples that {\em do} participate in the training at epoch t but with a wrong label. An effective SSL model minimizes both metrics: a low mask rate indicates that the model has access to more unlabeled examples during training (otherwise a low percentage and less diverse set of unlabeled examples are seen during training) while low impurity indicates that the pseudo-labels of these examples are of high quality. 
Note that we can compute impurity on these two datasets because our unlabeled data comes from the labeled training set of each of these datasets (thus we compare the pseudo-labels against the gold labels of each dataset).

As can be seen from the figures, FixMatch has a significantly larger mask rate due to the rigid confidence threshold set to a high value of $0.95$. In contrast, FlexMatch lowers the mask rate by $5\%$ with the introduction of flexible thresholds, but has a much higher impurity compared with FixMatch. Notably, our MarginMatch has only a slightly higher mask rate compared with FlexMatch and at the same time achieves a much lower impurity than FlexMatch and even FixMatch despite that FixMatch employs a very high confidence threshold. These results show that MarginMatch that enforces an additional measure for pseudo-labeled data quality maintains a low mask rate without compromising the quality of the pseudo-labels (i.e., low mask rate and low impurity). 

\subsection{Anecdotal Evidence}

\begin{figure} \centering
\begin{tikzpicture}[]   
     \node (fig1) at (1,1)
      {\begin{tikzpicture}[scale=0.45]
\begin{axis}[
        title=`Bird` and `Plane` APM and APM threshold,
        xlabel=$\#Epochs$,
        ylabel=$APM$,
        xmin=0, xmax=9000,
        ymin=-1.1, ymax=1.5,
        xtick={0,1000,...,9000},
        xticklabels={0, 1K, 2K, 3K, 4K, 5K, 6K, 7K, 8K,9K},
        ytick={-1, -0.5, 0, 0.5, 1.0},
        grid,
        every axis plot/.append style={ultra thick},
        every axis legend/.append style={at={(0.5,0.27)}},
        ]
\addplot[smooth,mark=*,blue,dashed] plot coordinates {
    (0,-0.3)
    (1000,-0.05)
    (2000, 0.14)
    (3000,-0.19)
    (4000, 0.27)
    (5000,-0.10)
    (6000,-0.34)
    (7000,-0.54)
    (8000,-0.49)
    (9000,-0.42)
};
\addlegendentry{`Plane` APM}

\addplot[smooth,mark=x,red,dash pattern=on 1pt off 3pt on 3pt off 3pt] plot coordinates {
    (0,-0.1)
    (1000,0.14)
    (2000,-0.17)
    (3000, 0.06)
    (4000, -0.23)
    (5000, 0.43)
    (6000, 0.79)
    (7000, 1.23)
    (8000, 2.28)
    (9000,3.45)
};
\addlegendentry{`Bird` APM}

\addplot[smooth,mark=x,black] plot coordinates {
    (0,0.24)
    (1000,0.27)
    (2000,0.54)
    (3000,0.52)
    (4000,0.51)
    (5000,0.54)
    (6000,0.57)
    (7000,0.55)
    (8000,0.58)
    (9000,0.55)
};
\addlegendentry{APM Threshold}

\end{axis}
\end{tikzpicture}};

    \node (fig2) at (5,1)
      {\begin{tikzpicture}[scale=0.45]
\begin{axis}[
        title=`Bird` and `Plane` APM and APM threshold,
        xlabel=$\#Epochs$,
        ylabel=$APM$,
        xmin=0, xmax=9000,
        ymin=-1.1, ymax=1.5,
        xtick={0,1000,...,9000},
        xticklabels={0, 1K, 2K, 3K, 4K, 5K, 6K, 7K, 8K,9K},
        ytick={-1, -0.5, 0, 0.5, 1.0},
        grid,
        every axis plot/.append style={ultra thick},
        every axis legend/.append style={at={(0.5,0.27)}},
        ]
\addplot[smooth,mark=*,blue,dashed] plot coordinates {
    (0,-0.3)
    (1000,-0.12)
    (2000,-0.23)
    (3000,-0.15)
    (4000,-0.14)
    (5000,-0.12)
    (6000,-0.11)
    (7000,-0.13)
    (8000,-0.14)
    (9000,-0.17)
};
\addlegendentry{`Plane` APM}

\addplot[smooth,mark=x,red,dash pattern=on 1pt off 3pt on 3pt off 3pt] plot coordinates {
    (0,-0.1)
    (1000,-0.15)
    (2000,0.04)
    (3000,0.08)
    (4000,0.26)
    (5000,0.74)
    (6000,1.23)
    (7000,1.42)
    (8000,2.65)
    (9000,3.17)
};
\addlegendentry{`Bird` APM}

\addplot[smooth,mark=x,black] plot coordinates {
    (0,0.24)
    (1000,0.27)
    (2000,0.54)
    (3000,0.52)
    (4000,0.51)
    (5000,0.54)
    (6000,0.57)
    (7000,0.55)
    (8000,0.58)
    (9000,0.55)
};
\addlegendentry{APM Threshold}

\end{axis}
\end{tikzpicture}};

    \node (fig3) at (1,5)
      {\begin{tikzpicture}[scale=0.45]
\begin{axis}[
        title=`Bird` and `Plane` confidence and confidence thresholds,
        xlabel=$\#Epochs$,
        ylabel=$Confidence$,
        xmin=0, xmax=9000,
        ymin=0, ymax=1.4,
        xtick={0,1000,...,9000},
        xticklabels={0, 1K, 2K, 3K, 4K, 5K, 6K, 7K, 8K,9K},
        ytick={0.25, 0.5, 0.75, 1.0},
        grid,
        every axis plot/.append style={ultra thick}
        ]
\addplot[smooth,mark=*,blue,dashed] plot coordinates {
    (0,0.1)
    (1000,0.23)
    (2000,0.55)
    (3000,0.56)
    (4000,0.72)
    (5000,0.83)
    (6000,0.82)
    (7000,0.81)
    (8000,0.88)
    (9000,0.93)
};
\addlegendentry{`Plane` Confidence}

\addplot[smooth,mark=x,red,dash pattern=on 1pt off 3pt on 3pt off 3pt] plot coordinates {
    (0,0.1)
    (1000,0.19)
    (2000,0.23)
    (3000,0.18)
    (4000,0.16)
    (5000,0.17)
    (6000,0.1)
    (7000,0.08)
    (8000,0.06)
    (9000,0.07)
};
\addlegendentry{`Bird` Confidence}

\addplot[smooth,mark=x,black] plot coordinates {
    (0,0.95)
    (1000,0.56)
    (2000,0.56)
    (3000,0.57)
    (4000,0.53)
    (5000,0.65)
    (6000,0.62)
    (7000,0.74)
    (8000,0.83)
    (9000,0.86)
};
\addlegendentry{`Bird` Confidence Thrs}
dash pattern=on 1pt off 3pt on 3pt off 3pt,

\addplot[smooth,mark=*,cyan] plot coordinates {
    (0,0.95)
    (1000,0.58)
    (2000,0.55)
    (3000,0.61)
    (4000,0.62)
    (5000,0.63)
    (6000,0.64)
    (7000,0.71)
    (8000,0.80)
    (9000,0.82)
};
\addlegendentry{`Plane` Confidence Thrs}
dash pattern=on 3pt off 2pt on 2pt off 3pt,

\end{axis}
\end{tikzpicture}};

 \node (fig4) at (5,5)
      {\begin{tikzpicture}[scale=0.45]
\begin{axis}[
        title=`Bird` and `Plane` confidence and confidence thresholds,
        xlabel=$\#Epochs$,
        ylabel=$Confidence$,
        xmin=0, xmax=9000,
        ymin=0, ymax=1.4,
        xtick={0,1000,...,9000},
        xticklabels={0, 1K, 2K, 3K, 4K, 5K, 6K, 7K, 8K,9K},
        ytick={0.25, 0.5, 0.75, 1.0},
        grid,
        every axis plot/.append style={ultra thick}
        ]
\addplot[smooth,mark=*,blue,dashed] plot coordinates {
    (0,0.3)
    (1000,0.35)
    (2000,0.34)
    (3000,0.32)
    (4000,0.35)
    (5000,0.25)
    (6000,0.3)
    (7000,0.21)
    (8000,0.18)
    (9000,0.17)
};
\addlegendentry{`Plane` Confidence}

\addplot[smooth,mark=x,red,dash pattern=on 1pt off 3pt on 3pt off 3pt] plot coordinates {
    (0,0.1)
    (1000,0.2)
    (2000,0.24)
    (3000,0.321)
    (4000,0.40)
    (5000,0.60)
    (6000,0.68)
    (7000,0.73)
    (8000,0.77)
    (9000,0.85)
};
\addlegendentry{`Bird` Confidence}

\addplot[smooth,mark=x,black] plot coordinates {
    (0,0.95)
    (1000,0.56)
    (2000,0.56)
    (3000,0.57)
    (4000,0.53)
    (5000,0.65)
    (6000,0.62)
    (7000,0.74)
    (8000,0.83)
    (9000,0.86)
};
\addlegendentry{`Bird` Confidence Thrs}
dash pattern=on 1pt off 3pt on 3pt off 3pt,

\addplot[smooth,mark=*,cyan] plot coordinates {
    (0,0.95)
    (1000,0.58)
    (2000,0.55)
    (3000,0.61)
    (4000,0.62)
    (5000,0.63)
    (6000,0.64)
    (7000,0.71)
    (8000,0.80)
    (9000,0.82)
};
\addlegendentry{`Plane` Confidence Thrs}
dash pattern=on 3pt off 2pt on 2pt off 3pt,

\end{axis}
\end{tikzpicture}};
\node[text width=1cm,rotate=90] at (-1.2,4.8){FlexMatch};
\node[text width=1cm,rotate=90] at (-1.2,0.8){MarginMatch};
     \node (fig4) at (4.1,5.9)
      {\includegraphics[scale=0.7]{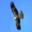}};   
     \node (fig6) at (4.17,1.9)
      {\includegraphics[scale=0.7]{2438.jpg}}; 
     \node (fig5) at (0.1,5.9)
      {\includegraphics[scale=0.7]{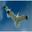}};  
     \node (fig7) at (0.17,1.9)
      {\includegraphics[scale=0.7]{1322.jpg}};
\end{tikzpicture}
\caption{Confidence thresholding vs. APM Thresholding on two images from the CIFAR-10 dataset.}
\label{motivation_conf_vs_apm}
\vspace{-6mm}
\end{figure}

\noindent
We show in Figure \ref{motivation_conf_vs_apm} anecdotal evidence of the effectiveness of MarginMatch. To this end, we extract two \emph{bird} images 
from our unlabeled portion of CIFAR-10 \cite{krizhevsky2009learning} of various learning difficulties that resemble characteristics of {\em plane} images (e.g., the background). The top part of the figure illustrates the progression over the training iterations of the confidence 
and the confidence thresholds of FlexMatch for the classes \emph{bird} and \emph{plane}, whereas the bottom part of the figure illustrates the progression of the APM threshold of MarginMatch along with its APMs of \emph{bird} and \emph{plane} classes over the training iterations. In the rightmost image, for MarginMatch we can observe that the APM of the {\em bird} class becomes stronger and stronger as the training progresses and eventually exceeds the APM threshold of MarginMatch, and hence, the image is included in the training until the end with the correct {\em bird} label. Interestingly, for the same image, in FlexMatch the confidence for the {\em bird} class is very close to the {\em bird} confidence threshold and eventually falls below this threshold and exits the training set. 

In contrast, the leftmost image is significantly more challenging than the rightmost image since it is more similar to 
\emph{plane} images, which makes it an easily confusable example. Here, we observe that the confidence of FlexMatch exceeds the flexible threshold with the incorrect argmax class \emph{plane} starting from iteration $3000$. Moreover, Flexmatch continues to use this image with the wrong \emph{plane} label for the remaining of the training process. 
Critically, in MarginMatch the APM value for the \emph{plane} class does not exceed the APM threshold, and the model eventually learns to classify this image correctly and includes it in training with the correct \emph{bird} pseudo-label.

\section{Related Work}

\noindent
\looseness=-1
Here, we focus on various SSL approaches that our MarginMatch directly builds upon although there are other SSL techniques that are not presented in this review, such as approaches based on generative models \cite{1640745,NIPS2007_4b6538a4}, graph-based approaches \cite{10.5555/645528.657646,10.5555/3041838.3041875} and robust SSL \cite{Park2021OpenCoSCS,NEURIPS2021_da11e8cd} (see Appendix C for a comparison between MarginMatch and Robust SSL approaches). 

\textbf{Self-training} \cite{xie2020self,1053799,mcclosky-etal-2006-effective,4129456} is a popular SSL method where the predictions of a model on unlabeled data are used as artificial labels to train against. Noisy student training \cite{xie2020self}  is a popular self-training approach that also leverages knowledge distillation \cite{44873} and iteratively jointly trains two models in a teacher-student framework. Noisy student uses a larger model size and noised inputs, exposing the student to more difficult learning environments, leading to an increased performance compared to the teacher.

\textbf{Pseudo-labeling} is a variant of self-training where these predictions are sharpened to obtain hard labels \cite{lee2013pseudo}. The use of hard labels can be seen as a means of entropy minimization \cite{NIPS2004_96f2b50b} and nowadays is a valuable component in most successful SSL approaches \cite{berthelot2019mixmatch,berthelot2019remixmatch,
NEURIPS2021_995693c1}. These hard labels are usually used along a confidence threshold, where unconfident unlabeled examples are completely disregarded (e.g., \cite{berthelot2019mixmatch}) to avoid using noisy pseudo-labels. Recently, approaches such as Curriculum Labeling (CL) \cite{CascanteBonilla2021CurriculumLR} or FlexMatch \cite{NEURIPS2021_995693c1} started to explore curriculum learning in the SSL context. CL proposes a self-pacing strategy of identifying easy and hard examples to ensure that the model first uses easy and progressively moves towards harder examples. Similarly, MarginMatch uses curriculum learning and pseudo-labeling, but the focus of our approach is placed on producing better thresholds for assessing the quality of pseudo-labels.

\looseness=-1
\textbf{Consistency regularization} \cite{NIPS2014_66be31e4} is a method that applies random perturbations when generating the artificial label, such as data augmentation \cite{berthelot2019mixmatch,berthelot2019remixmatch,sohn2020fixmatch}, dropout \cite{sajjadi2016regularization}, or adversarial perturbations \cite{miyato2016adversarial}. Current state-of-the-art approaches \cite{sohn2020fixmatch,NEURIPS2021_995693c1} 
exploit a combination of weak and strong data augmentations, which were shown to be extremely beneficial in SSL. The most popular strong augmentations used in the SSL literature are RandAugment \cite{cubuk2020randaugment} and CTAugment \cite{berthelot2019remixmatch}. The approaches based on these methods first generate a hard label using pseudo-labeling on a weakly augmented image (i.e., using a low noise transformation such as a flip-and-shift augmentation), then optimize the predictions of the model on a strongly augmented version of the same image towards this hard label. 
Similar to these approaches, MarginMatch uses the same combination of weak and strong data augmentations. 

\section{Conclusion}

In this paper, we proposed a novel semi-supervised learning method that improves the pseudo-label quality using training dynamics. Our new method is lightweight and achieves state-of-the-art performance on four computer vision SSL datasets in low data regimes and on two large-scale benchmarks. MarginMatch takes into consideration not only a flexible confidence threshold to account for the difficulty of each class, but also a measure of quality for each unlabeled example using training dynamics. In addition, MarginMatch is a general approach that can be leveraged in most SSL frameworks and 
we hope that it can attract future research in analyzing the effectiveness of SSL approaches focused on data quality. As future work, we aim to further explore our method in settings when there is a mismatch between the labeled and unlabeled data distributions (i.e., making use of out-of-domain unlabeled data).

\newpage

{\small
\bibliographystyle{ieee_fullname}
\bibliography{egbib}

\begin{thebibliography}{10}\itemsep=-1pt

\bibitem{DBLP:journals/corr/abs-1908-02983}
Eric Arazo, Diego Ortego, Paul Albert, Noel~E. O'Connor, and Kevin McGuinness.
\newblock Pseudo-labeling and confirmation bias in deep semi-supervised
  learning.
\newblock {\em CoRR}, abs/1908.02983, 2019.

\bibitem{NIPS2014_66be31e4}
Philip Bachman, Ouais Alsharif, and Doina Precup.
\newblock Learning with pseudo-ensembles.
\newblock In Z. Ghahramani, M. Welling, C. Cortes, N. Lawrence, and K.Q.
  Weinberger, editors, {\em Advances in Neural Information Processing Systems},
  volume~27. Curran Associates, Inc., 2014.

\bibitem{DBLP:journals/corr/BartlettFT17}
Peter~L. Bartlett, Dylan~J. Foster, and Matus Telgarsky.
\newblock Spectrally-normalized margin bounds for neural networks.
\newblock {\em CoRR}, abs/1706.08498, 2017.

\bibitem{berthelot2019remixmatch}
David Berthelot, Nicholas Carlini, Ekin~D Cubuk, Alex Kurakin, Kihyuk Sohn, Han
  Zhang, and Colin Raffel.
\newblock Remixmatch: Semi-supervised learning with distribution alignment and
  augmentation anchoring.
\newblock {\em arXiv preprint arXiv:1911.09785}, 2019.

\bibitem{berthelot2019mixmatch}
David Berthelot, Nicholas Carlini, Ian Goodfellow, Nicolas Papernot, Avital
  Oliver, and Colin Raffel.
\newblock Mixmatch: A holistic approach to semi-supervised learning.
\newblock {\em arXiv preprint arXiv:1905.02249}, 2019.

\bibitem{CascanteBonilla2021CurriculumLR}
Paola Cascante-Bonilla, Fuwen Tan, Yanjun Qi, and Vicente Ordonez.
\newblock Curriculum labeling: Revisiting pseudo-labeling for semi-supervised
  learning.
\newblock In {\em AAAI}, 2021.

\bibitem{chen2020simple}
Ting Chen, Simon Kornblith, Mohammad Norouzi, and Geoffrey Hinton.
\newblock A simple framework for contrastive learning of visual
  representations.
\newblock In {\em International conference on machine learning}, pages
  1597--1607. PMLR, 2020.

\bibitem{coates2011analysis}
Adam Coates, Andrew Ng, and Honglak Lee.
\newblock An analysis of single-layer networks in unsupervised feature
  learning.
\newblock In {\em Proceedings of the fourteenth international conference on
  artificial intelligence and statistics}, pages 215--223. JMLR Workshop and
  Conference Proceedings, 2011.

\bibitem{cubuk2020randaugment}
Ekin~D Cubuk, Barret Zoph, Jonathon Shlens, and Quoc~V Le.
\newblock Randaugment: Practical automated data augmentation with a reduced
  search space.
\newblock In {\em Proceedings of the IEEE/CVF Conference on Computer Vision and
  Pattern Recognition Workshops}, pages 702--703, 2020.

\bibitem{5206848}
Jia Deng, Wei Dong, Richard Socher, Li-Jia Li, Kai Li, and Li Fei-Fei.
\newblock Imagenet: A large-scale hierarchical image database.
\newblock In {\em 2009 IEEE Conference on Computer Vision and Pattern
  Recognition}, pages 248--255, 2009.

\bibitem{47365}
Gamaleldin~Fathy Elsayed, Dilip Krishnan, Hossein Mobahi, Kevin Regan, and Samy
  Bengio.
\newblock Large margin deep networks for classification.
\newblock 2018.

\bibitem{NIPS2004_96f2b50b}
Yves Grandvalet and Yoshua Bengio.
\newblock Semi-supervised learning by entropy minimization.
\newblock In L. Saul, Y. Weiss, and L. Bottou, editors, {\em Advances in Neural
  Information Processing Systems}, volume~17. MIT Press, 2004.

\bibitem{pmlr-v70-guo17a}
Chuan Guo, Geoff Pleiss, Yu Sun, and Kilian~Q. Weinberger.
\newblock On calibration of modern neural networks.
\newblock In Doina Precup and Yee~Whye Teh, editors, {\em Proceedings of the
  34th International Conference on Machine Learning}, volume~70 of {\em
  Proceedings of Machine Learning Research}, pages 1321--1330. PMLR, 06--11 Aug
  2017.

\bibitem{he2016deep}
Kaiming He, Xiangyu Zhang, Shaoqing Ren, and Jian Sun.
\newblock Deep residual learning for image recognition.
\newblock In {\em Proceedings of the IEEE conference on computer vision and
  pattern recognition}, pages 770--778, 2016.

\bibitem{hestness2017deep}
Joel Hestness, Sharan Narang, Newsha Ardalani, Gregory Diamos, Heewoo Jun,
  Hassan Kianinejad, Md Patwary, Mostofa Ali, Yang Yang, and Yanqi Zhou.
\newblock Deep learning scaling is predictable, empirically.
\newblock {\em arXiv preprint arXiv:1712.00409}, 2017.

\bibitem{44873}
Geoffrey Hinton, Oriol Vinyals, and Jeffrey Dean.
\newblock Distilling the knowledge in a neural network.
\newblock In {\em NIPS Deep Learning and Representation Learning Workshop},
  2015.

\bibitem{NIPS2007_4b6538a4}
Geoffrey~E Hinton and Russ~R Salakhutdinov.
\newblock Using deep belief nets to learn covariance kernels for gaussian
  processes.
\newblock In J. Platt, D. Koller, Y. Singer, and S. Roweis, editors, {\em
  Advances in Neural Information Processing Systems}, volume~20. Curran
  Associates, Inc., 2007.

\bibitem{https://doi.org/10.48550/arxiv.1810.00113}
Yiding Jiang, Dilip Krishnan, Hossein Mobahi, and Samy Bengio.
\newblock Predicting the generalization gap in deep networks with margin
  distributions, 2018.

\bibitem{10.5555/645528.657646}
Thorsten Joachims.
\newblock Transductive inference for text classification using support vector
  machines.
\newblock In {\em Proceedings of the Sixteenth International Conference on
  Machine Learning}, ICML '99, page 200–209, San Francisco, CA, USA, 1999.
  Morgan Kaufmann Publishers Inc.

\bibitem{10.5555/3041838.3041875}
Thorsten Joachims.
\newblock Transductive learning via spectral graph partitioning.
\newblock In {\em Proceedings of the Twentieth International Conference on
  International Conference on Machine Learning}, ICML'03, page 290–297. AAAI
  Press, 2003.

\bibitem{krizhevsky2009learning}
Alex Krizhevsky, Geoffrey Hinton, et~al.
\newblock Learning multiple layers of features from tiny images.
\newblock 2009.

\bibitem{10.5555/2999134.2999257}
Alex Krizhevsky, Ilya Sutskever, and Geoffrey~E. Hinton.
\newblock Imagenet classification with deep convolutional neural networks.
\newblock In {\em Proceedings of the 25th International Conference on Neural
  Information Processing Systems - Volume 1}, NIPS'12, page 1097–1105, Red
  Hook, NY, USA, 2012. Curran Associates Inc.

\bibitem{laine2016temporal}
Samuli Laine and Timo Aila.
\newblock Temporal ensembling for semi-supervised learning.
\newblock In {\em ICLR (Poster)}. OpenReview.net, 2017.

\bibitem{1640745}
J.A. Lasserre, C.M. Bishop, and T.P. Minka.
\newblock Principled hybrids of generative and discriminative models.
\newblock In {\em 2006 IEEE Computer Society Conference on Computer Vision and
  Pattern Recognition (CVPR'06)}, volume~1, pages 87--94, 2006.

\bibitem{lee2013pseudo}
Dong-Hyun Lee et~al.
\newblock Pseudo-label: The simple and efficient semi-supervised learning
  method for deep neural networks.
\newblock In {\em Workshop on challenges in representation learning, ICML},
  volume~3, page 896, 2013.

\bibitem{DBLP:journals/corr/abs-1708-02862}
Wen Li, Limin Wang, Wei Li, Eirikur Agustsson, and Luc~Van Gool.
\newblock Webvision database: Visual learning and understanding from web data.
\newblock {\em CoRR}, abs/1708.02862, 2017.

\bibitem{mahajan2018exploring}
Dhruv Mahajan, Ross Girshick, Vignesh Ramanathan, Kaiming He, Manohar Paluri,
  Yixuan Li, Ashwin Bharambe, and Laurens Van Der~Maaten.
\newblock Exploring the limits of weakly supervised pretraining.
\newblock In {\em Proceedings of the European conference on computer vision
  (ECCV)}, pages 181--196, 2018.

\bibitem{mcclosky-etal-2006-effective}
David McClosky, Eugene Charniak, and Mark Johnson.
\newblock Effective self-training for parsing.
\newblock In {\em Proceedings of the Human Language Technology Conference of
  the {NAACL}, Main Conference}, pages 152--159, New York City, USA, June 2006.
  Association for Computational Linguistics.

\bibitem{miyato2016adversarial}
Takeru Miyato, Andrew~M Dai, and Ian Goodfellow.
\newblock Adversarial training methods for semi-supervised text classification.
\newblock {\em arXiv preprint arXiv:1605.07725}, 2016.

\bibitem{miyato2018virtual}
Takeru Miyato, Shin-ichi Maeda, Masanori Koyama, and Shin Ishii.
\newblock Virtual adversarial training: a regularization method for supervised
  and semi-supervised learning.
\newblock {\em IEEE transactions on pattern analysis and machine intelligence},
  41(8):1979--1993, 2018.

\bibitem{netzer2011reading}
Yuval Netzer, Tao Wang, Adam Coates, Alessandro Bissacco, Bo Wu, and Andrew~Y
  Ng.
\newblock Reading digits in natural images with unsupervised feature learning.
\newblock 2011.

\bibitem{Park2021OpenCoSCS}
Jongjin Park, Sukmin Yun, Jongheon Jeong, and Jinwoo Shin.
\newblock Opencos: Contrastive semi-supervised learning for handling open-set
  unlabeled data.
\newblock In {\em ECCV Workshops}, 2021.

\bibitem{NEURIPS2020_c6102b37}
Geoff Pleiss, Tianyi Zhang, Ethan Elenberg, and Kilian~Q Weinberger.
\newblock Identifying mislabeled data using the area under the margin ranking.
\newblock In H. Larochelle, M. Ranzato, R. Hadsell, M.~F. Balcan, and H. Lin,
  editors, {\em Advances in Neural Information Processing Systems}, volume~33,
  pages 17044--17056. Curran Associates, Inc., 2020.

\bibitem{radford2019language}
Alec Radford, Jeffrey Wu, Rewon Child, David Luan, Dario Amodei, Ilya
  Sutskever, et~al.
\newblock Language models are unsupervised multitask learners.
\newblock {\em OpenAI blog}, 1(8):9, 2019.

\bibitem{raffel2019exploring}
Colin Raffel, Noam Shazeer, Adam Roberts, Katherine Lee, Sharan Narang, Michael
  Matena, Yanqi Zhou, Wei Li, and Peter~J Liu.
\newblock Exploring the limits of transfer learning with a unified text-to-text
  transformer.
\newblock {\em arXiv preprint arXiv:1910.10683}, 2019.

\bibitem{4129456}
Chuck Rosenberg, Martial Hebert, and Henry Schneiderman.
\newblock Semi-supervised self-training of object detection models.
\newblock In {\em 2005 Seventh IEEE Workshops on Applications of Computer
  Vision (WACV/MOTION'05) - Volume 1}, volume~1, pages 29--36, 2005.

\bibitem{NEURIPS2021_da11e8cd}
Kuniaki Saito, Donghyun Kim, and Kate Saenko.
\newblock Openmatch: Open-set semi-supervised learning with open-set
  consistency regularization.
\newblock In M. Ranzato, A. Beygelzimer, Y. Dauphin, P.S. Liang, and J.~Wortman
  Vaughan, editors, {\em Advances in Neural Information Processing Systems},
  volume~34, pages 25956--25967. Curran Associates, Inc., 2021.

\bibitem{sajjadi2016mutual}
Mehdi Sajjadi, Mehran Javanmardi, and Tolga Tasdizen.
\newblock Mutual exclusivity loss for semi-supervised deep learning.
\newblock In {\em 2016 IEEE International Conference on Image Processing
  (ICIP)}, pages 1908--1912. IEEE, 2016.

\bibitem{sajjadi2016regularization}
Mehdi Sajjadi, Mehran Javanmardi, and Tolga Tasdizen.
\newblock Regularization with stochastic transformations and perturbations for
  deep semi-supervised learning.
\newblock {\em Advances in neural information processing systems},
  29:1163--1171, 2016.

\bibitem{1053799}
H. Scudder.
\newblock Probability of error of some adaptive pattern-recognition machines.
\newblock {\em IEEE Transactions on Information Theory}, 11(3):363--371, 1965.

\bibitem{sohn2020fixmatch}
Kihyuk Sohn, David Berthelot, Nicholas Carlini, Zizhao Zhang, Han Zhang,
  Colin~A Raffel, Ekin~Dogus Cubuk, Alexey Kurakin, and Chun-Liang Li.
\newblock Fixmatch: Simplifying semi-supervised learning with consistency and
  confidence.
\newblock In H. Larochelle, M. Ranzato, R. Hadsell, M.~F. Balcan, and H. Lin,
  editors, {\em Advances in Neural Information Processing Systems}, volume~33,
  pages 596--608. Curran Associates, Inc., 2020.

\bibitem{szegedy2015going}
Christian Szegedy, Wei Liu, Yangqing Jia, Pierre Sermanet, Scott Reed, Dragomir
  Anguelov, Dumitru Erhan, Vincent Vanhoucke, and Andrew Rabinovich.
\newblock Going deeper with convolutions.
\newblock In {\em Proceedings of the IEEE conference on computer vision and
  pattern recognition}, pages 1--9, 2015.

\bibitem{tan2019efficientnet}
Mingxing Tan and Quoc Le.
\newblock Efficientnet: Rethinking model scaling for convolutional neural
  networks.
\newblock In {\em International Conference on Machine Learning}, pages
  6105--6114. PMLR, 2019.

\bibitem{tarvainen2017mean}
Antti Tarvainen and Harri Valpola.
\newblock Mean teachers are better role models: Weight-averaged consistency
  targets improve semi-supervised deep learning results.
\newblock In {\em Advances in neural information processing systems}, pages
  1195--1204, 2017.

\bibitem{verma2021interpolation}
Vikas Verma, Kenji Kawaguchi, Alex Lamb, Juho Kannala, Arno Solin, Yoshua
  Bengio, and David Lopez-Paz.
\newblock Interpolation consistency training for semi-supervised learning.
\newblock {\em Neural Networks}, 2021.

\bibitem{xie2019unsupervised}
Qizhe Xie, Zihang Dai, Eduard Hovy, Thang Luong, and Quoc Le.
\newblock Unsupervised data augmentation for consistency training.
\newblock In H. Larochelle, M. Ranzato, R. Hadsell, M.~F. Balcan, and H. Lin,
  editors, {\em Advances in Neural Information Processing Systems}, volume~33,
  pages 6256--6268. Curran Associates, Inc., 2020.

\bibitem{xie2020self}
Qizhe Xie, Minh-Thang Luong, Eduard Hovy, and Quoc~V Le.
\newblock Self-training with noisy student improves imagenet classification.
\newblock In {\em Proceedings of the IEEE/CVF Conference on Computer Vision and
  Pattern Recognition}, pages 10687--10698, 2020.

\bibitem{zagoruyko2017wide}
Sergey Zagoruyko and Nikos Komodakis.
\newblock Wide residual networks, 2017.

\bibitem{NEURIPS2021_995693c1}
Bowen Zhang, Yidong Wang, Wenxin Hou, HAO WU, Jindong Wang, Manabu Okumura, and
  Takahiro Shinozaki.
\newblock Flexmatch: Boosting semi-supervised learning with curriculum pseudo
  labeling.
\newblock In M. Ranzato, A. Beygelzimer, Y. Dauphin, P.S. Liang, and J.~Wortman
  Vaughan, editors, {\em Advances in Neural Information Processing Systems},
  volume~34, pages 18408--18419. Curran Associates, Inc., 2021.

\bibitem{DBLP:journals/corr/ZhangBHRV16}
Chiyuan Zhang, Samy Bengio, Moritz Hardt, Benjamin Recht, and Oriol Vinyals.
\newblock Understanding deep learning requires rethinking generalization.
\newblock {\em CoRR}, abs/1611.03530, 2016.

\end{thebibliography}
}

\end{document}


\title{Supplemetary Material}

\author{Tiberiu Sosea \quad\quad\quad  Cornelia Caragea \\
University of Illinois Chicago\\
{\small {\color{blue}{\tt tsosea2@uic.edu}} \quad\quad\quad {\color{blue}{\tt cornelia@uic.edu}}}
}
\maketitle

\appendix

\section{Additional details of threshold Class C+1}

\noindent
We train the model to output class C+1 on threshold examples. We aim to mimic the margin of mislabeled examples (e.g., an image of a bird labeled as a cat) and intentionally mislabel a random set of unlabeled examples with a new virtual class.  By moving these examples in this virtual class $C+1$, we ensure that this class likely contains examples from each of the $C$ classes which will be mislabeled with the inexistent class C+1. For an unlabeled example that now belongs to this virtual class, the model tends to label it in its correct class $c \in C$ through generalization (based on other images that correctly belong to class $c$), so the difference between the virtual class $C+1$ and the argmax (i.e., probably of $c$) will be negative, which is similar to the margins of mislabeled examples. 

\section{Insights into why EMA Margin outperforms EMA Confidence and Entropy}

\noindent
Let $x_{i}$ be an unlabeled example where our model $M$ fluctuates significantly between iterations. Entropy captures uncertainty through the class probability distribution, but it does not consider the argmax of the prediction. If the class distribution of $x_{i}$ has one peak but the peak fluctuates between classes from one iteration to another, the average entropy will be low and hence, $x_{i}$ will be incorrectly included in the training. Confidence measures the magnitude of the probability of predicting a class $c$ and does not penalize too much cases when the distribution has two peaks with fairly similar probabilities (with the other classes having probabilities close to $0$). However, unlike entropy, it penalizes changes in argmax across iterations. The margin solves both drawbacks. Let $t$ be the current iteration and $c$ be the argmax at this iteration (hence the \emph{ground truth}; please see line 313 in the main paper). If at some iteration $t'<t$ the predicted class is $c'$, $c'\neq c$, then the difference between the logits corresponding to $c$ and $c'$ is negative since $c'$ corresponds to the argmax at $t'$. Thus, frequent fluctuations up to iteration $t$ will yield likely negative averaged marings even if the margin for the \emph{ground truth} class $c$ will be large at iteration $t$ but small at iteration $t'$.

\section{Comparison with Robust SSL Approaches}

\begin{figure}
\centering
\begin{tabular}{l|c}
\toprule
Method &  Error Rate\\
\midrule
MarginMatch & $25.37$ \\
FixMatch & $38.43$ \\
FlexMatch & $29.56$ \\
OpenCos & $31.53$ \\
OpenMatch & $35.12$ \\
\bottomrule
\end{tabular}
\caption{Error rates on STL-10 with $4$ examples per class.}
\label{robust_ssl}
\end{figure}

\noindent
We carried out an additional experiment to compare our method with OpenMatch \cite{NEURIPS2021_da11e8cd}  and OpenCos \cite{Park2021OpenCoSCS}. Since both these approaches address the case where the unlabeled data contains a different label space from the labeled data, we only include results on STL-10 using $4$ examples per class, since it is the only dataset with these properties. We show these results in Table \ref{robust_ssl} where we observe that both FlexMatch and our MarginMatch outperform both methods considerably.

{\small
\bibliographystyle{ieee_fullname}
\bibliography{egbib}
}